%% file: main.tex
\documentclass[letterpaper]{article} 
\usepackage{aaai25}  
\usepackage{times}  
\usepackage{helvet}  
\usepackage{courier}  
\usepackage[hyphens]{url}  
\usepackage{graphicx} 
\usepackage{natbib}  
\usepackage{caption} 
\usepackage{pgfplots}
\usepackage{subcaption}
\usepackage{pgf-pie}
\usepackage{tikz}
\pgfplotsset{compat=1.17}
\usepackage{amssymb}
\usepackage{pifont}
\usepackage{algorithm}
\usepackage{algorithmic}
\usepackage{booktabs}

\usepackage{newfloat}
\usepackage{listings}
\DeclareCaptionStyle{ruled}{labelfont=normalfont,labelsep=colon,strut=off} 
\floatstyle{ruled}
\newfloat{listing}{tb}{lst}{}
\floatname{listing}{Listing}
\title{Fairness in Federated Learning: Fairness for Whom?}

\title{Fairness in Federated Learning: Fairness for Whom?}
\author {
    Afaf Taik\textsuperscript{\rm 1,\rm 2},
    Khaoula Chehbouni\textsuperscript{\rm 1,\rm 3},
    Golnoosh Farnadi\textsuperscript{\rm 1, \rm 2, \rm 3}
}
\affiliations {
    \textsuperscript{\rm 1}Mila - Quebec AI Institute\\
    \textsuperscript{\rm 2}Université de Montréal\\
    \textsuperscript{\rm 3}McGill University\\
    afaf.taik@mila.quebec, khaoula.chehbouni@mila.quebec, farnadig@mila.quebec
}

\usepackage{bibentry}

\begin{document}

\definecolor{darkorchid}{rgb}{0.6, 0.2, 0.8}
\newcommand{\khaoula}[1]{{\color{darkorchid}{\small\bf\sf [Khaoula: #1]}}}
\newcommand{\afaf}[1]{{\color{teal}{\small\bf\sf [Afaf: #1]}}}
\newcommand{\clea}[1]{{\color{purple}{\small\bf\sf [Cléa: #1]}}}

\maketitle

\begin{abstract}
Fairness in federated learning (FL) has emerged as a rapidly growing area of research, with numerous works proposing formal definitions and algorithmic interventions. Yet, despite this technical progress, fairness in FL is often defined and evaluated in ways that abstract away from the sociotechnical contexts in which these systems are deployed. In this paper, we argue that existing approaches tend to optimize narrow system-level metrics --- such as performance parity or contribution-based rewards --- while overlooking how harms arise throughout the FL lifecycle and how they impact diverse stakeholders. We support this claim through a critical analysis of the literature, based on a systematic annotation of papers for their fairness definitions, design decisions, evaluation practices, and motivating use cases. Our analysis reveals five recurring pitfalls: (1) fairness framed solely through the lens of server–client architecture; (2) a mismatch between simulations and motivating use-cases and contexts; (3) definitions that conflate protecting the system with protecting its users; (4) interventions that target isolated stages of the lifecycle while neglecting upstream and downstream effects; and (5) a lack of multi-stakeholder alignment where multiple fairness definitions can be relevant at once. Building on these insights, we propose a harm-centered framework that links fairness definitions to concrete risks and stakeholder vulnerabilities. We conclude with recommendations for more holistic, context-aware, and accountable fairness research in FL.
\end{abstract}

\input{01_introduction_bis}

\input{02_background}
\input{03_failures}

\input{04_bias_framework}

\input{05_takeaways}

\input{06_limitations_idk}

\bibliography{aaai25}

\end{document}

%% file: 01_introduction_bis.tex
\section{Introduction}

Machine learning (ML) systems are increasingly deployed in high-stakes domains such as healthcare~\cite{hadi2023survey} and hiring~\cite{qin2018enhancing}, where their decisions carry significant social implications. Ensuring fairness in such systems is critical, yet difficult. While fairness in centralized ML has been widely studied~\cite{mitchell2021algorithmic, saxena2019fairness}, collaborative ML paradigms such as Federated Learning (FL) introduce new complexities to this endeavor~\cite{konevcny2016federated, kairouz2021advances}. In FL, model training is distributed across multiple clients (e.g., smartphones or institutions), each with their own local data and constraints. This setup brings unique challenges, including heterogeneity in data, compute, and context, as well as limited observability into client distributions or downstream impact.

To address these challenges, new fairness definitions have emerged specifically for FL~\cite{shi_towards_2024}, including participation fairness, performance fairness, group fairness, and collaborative fairness. Several recent surveys have documented these proposals~\cite{shi_towards_2024, vucinich_current_2023, annapareddy_fairness_2023, salazar_survey_2024, ude_survey_2023, rafi_fairness_2024, chen_privacy_2024, wang_linkage_2024, balbierer_multivocal_2024}. While this body of work provides valuable taxonomies and technical insight, it often treats fairness as a standalone optimization problem, isolated from the social context in which FL systems are deployed.

In this paper, we argue that much of the fairness in FL literature suffers from a critical abstraction error~\cite{selbst2019fairness}. Fairness is often defined and operationalized in terms of narrow system-level metrics, such as minimizing performance variance across clients or allocating rewards based on contribution, without attention to the structural inequalities, institutional dynamics, or real-world harms that shape FL participation and impact. These abstractions risk obscuring who is protected, what harms are being addressed, and whose interests are prioritized.

To examine these concerns, we adopt a harm-centered lens and conduct a systematic annotation of 121 papers on fairness in FL. We map fairness definitions, intervention points, and evaluation practices to the FL lifecycle—from problem formulation to deployment—and identify recurring gaps in how fairness is conceptualized and operationalized. Our findings reveal five core pitfalls: (i) a narrow focus on server–client architecture; (ii) a disconnect between fairness definitions and simulated use-cases; (iii) the conflation of system  interests' protection with user protection; (iv) a limited intervention scope for mitigation techniques; and (v) a lack of multi-stakeholder alignment where many fairness definitions could be relevant. 

We also highlight how fairness efforts in FL interact with other concerns such as privacy and robustness—sometimes introducing new harms, such as disproportionate privacy degradation for minorities~\cite{ling_fedfdp_2024}, or the misclassification of underrepresented clients as adversaries~\cite{touat_towards_2023}.

Furthermore, to address these gaps, we propose a harm-centered framework for guiding context-aware fair FL solutions. Our framework maps the FL lifecycle—spanning problem formulation, client selection, aggregation, evaluation, and deployment—against different sources of harm and stakeholder vulnerabilities. It encourages researchers to ask: Who is (or could be) harmed at this step? What kinds of errors or exclusions may occur? And how do these relate to existing fairness definitions? This approach offers a more grounded and context-aware lens for evaluating fairness claims and designing interventions.

Our contributions are twofold:
\begin{itemize}
    \item We provide a critical, annotated review of the fairness in FL literature, uncovering patterns in definitions, motivations, and lifecycle coverage.
    \item We propose a harm-centered framework for fairness in FL, connecting technical decisions to stakeholder impact, and offering actionable recommendations for more accountable, context-aware research.
\end{itemize}

%% file: 02_background.tex
\section{Background}

Federated learning (FL) is a decentralized ML paradigm in which a global model is trained collaboratively across multiple participants (e.g., smartphones, hospitals, or institutions), without exchanging raw data. Clients train the model locally and share updates (e.g., gradients or model weights) with a central server, which aggregates them to produce an updated model. This process repeats over multiple communication rounds \cite{kairouz2021advances}. Depending on the type of participants, we refer to the setting as cross-silo FL for institutional participants, and cross-device FL when clients are massively distributed IoT and mobile devices. This distinction helps move beyond the abstract client-server model by grouping use cases with similar resource constraints and aligning them with the types of stakeholders both operating the clients and affected by the resulting model—for example, mobile users both provide data and are affected by the model in cross-device FL, whereas in cross-silo FL, institutions (e.g., banks, hospitals) operate as the clients, but use their models to make decisions impacting their customers. 

While FL is often introduced as a privacy-preserving alternative to centralized ML, its distributed nature introduces new technical and ethical challenges. In particular, fairness in FL must be reconsidered in light of structural asymmetries between clients. These asymmetries are not incidental; they are deeply ingrained into the realities of FL deployment.
\subsection{Federated Learning Lifecycle}
Although implementations of FL vary across settings and applications, most systems follow a common lifecycle that structures the learning process and coordination between clients and the central server. The following outline represents a basic FL lifecycle. While researchers have proposed many variations---including hierarchical \cite{abad2020hierarchical} and clustered \cite{el2025survey} setups, and personalization training schemes \cite{tan2022towards}---these core steps remain the most widely adopted and studied in practice: 

\begin{enumerate}
    \item \textbf{Problem Formulation:}  
    The task to be solved is selected, formalized, and translated into an ML problem. This includes defining objectives, choosing input/output formats, loss functions, and defining the different constraints. 
    
    \item \textbf{Model Initialization:} An initial global model is defined—often using a standard architecture and optionally initialized with pretrained weights. This phase also includes setting key hyperparameters (e.g., learning rate, batch size), and establishing the stopping criterion (e.g., number of rounds or convergence threshold). 
    Then, in each iteration, referred to as communication round, the following steps 3 to 5 are repeated until the stopping criterion is met. 
    
    \item \textbf{Client Selection:}  
     During training, a subset of clients (e.g., institutions, mobile devices) is selected in each round based on varying criteria such as availability, resources, or data attributes. These choices affect which clients contribute to the model.

    \item \textbf{Local Training:}  
    Clients train the model locally on their private data. Key design decisions at this stage include number of local epochs, optimizer choices, and any personalization applied before sending updates.

    \item \textbf{Model Aggregation:}  
    The central server aggregates client updates (e.g., using weighted averaging). The aggregation strategy and timing (synchronous/asynchronous) influence which updates contribute and how they are valued.

    \item \textbf{Evaluation:}  
    The model is evaluated—either centrally, on a held-out dataset, or by clients locally. Evaluation choices affect which metrics are prioritized and accounted for.

    \item \textbf{Deployment and Incentives:}  
    The final model (or personalized variants) is deployed to clients or users. In some settings, reward mechanisms may be used to incentivize participation, often based on estimated contributions.

\end{enumerate}

In addition to this cycle, further mechanisms are often added to enhance privacy and guarantee adversarial robustness.





\subsection{Why Fairness in FL is Challenging}
\label{subsec:fairnesschallenges}


FL is fundamentally shaped by two structural properties of FL systems: \textit{heterogeneity} and \textit{scarcity}.

\textbf{Heterogeneity} refers to differences across clients along three critical dimensions \cite{10.1145/3617694.3623256}:
\begin{itemize}
    \item \textit{Data heterogeneity}: Local datasets may differ in size, label distribution, feature representation, or demographic composition.
    \item \textit{Resource heterogeneity}: Clients vary in computational capacity, network stability, and energy constraints.
    \item \textit{Contextual heterogeneity}: Clients operate under diverse legal, institutional, or social conditions that shape what data can be collected or shared.
\end{itemize}

\textbf{Scarcity}, meanwhile, refers to limited availability of data and compute resources—especially among clients serving minority populations or operating under infrastructural constraints. Scarcity amplifies the risk that some clients will be consistently underrepresented or excluded from the training process.

Together, heterogeneity and scarcity lead to two key limitations for FL: 
(i) \textit{Partial participation}: Not all clients participate equally or consistently, leading to sampling biases in model updates; (ii) \textit{Limited visibility}: The server typically lacks access to protected attributes, demographic breakdowns, or downstream outcomes—making it difficult to measure or enforce fairness during training.

\subsection{How Fairness is Defined in FL Research}

Across the literature, several fairness notions have been proposed for FL \cite{shi_towards_2024}, often adapted from centralized ML or cooperative systems. These include:

\begin{itemize}
     \item \textbf{Performance-centered fairness}: 
     The dominant definition of performance fairness found in the FL literature \cite{li_fair_2020,yan_new_2024,xu_achieving_2022,chu_rethinking_2024,cong_ada-ffl_2024} refers to achieving similar or comparable model performance across different clients or groups of clients. A model $\omega_1$ is considered fairer than model $\omega_2$ with respect to the participants if, for a metric of interest (e.g., accuracy, loss), it has a smaller variance across clients. This definition can also be applied when comparing multiple client groups. Two additional definitions were proposed in this category: 1) Rawlsian definition of fairness, where the objective is to optimize the performance of the worst performing clients or group of clients \cite{oksuz_boosting_2024,mohri2019agnostic}; and 2) 
individual fairness, where similar clients should have similar performance \cite{mashhadi_auditing_2022}.  

    \item \textbf{Group fairness-inspired definitions}: Group fairness in ML emphasizes that predictions should not disadvantage underrepresented or unprivileged groups based on sensitive attributes such as race or gender. 
    Building on these principles, researchers have sought to adapt classical group fairness criteria---such as equal opportunity, equalized odds, or demographic parity---to the federated learning setting, applying them both to global aggregates and to individual client models \cite{shi_towards_2024, salazar_survey_2024}. While these definitions are applicable to cross-device settings, they are mostly studied in the context of cross-silo collaborations, where institutions develop models that need to be fair to the populations they serve. 

    \item \textbf{Collaborative fairness}: Ensuring that clients receive rewards, utility, or model quality proportionally to their contributions. This definition of fairness is particularly relevant for cross-silo FL settings, where data owners often compete within the same market and face increased risks from malicious actors and free-riders. \citet{yu_fairness-aware_2020} define three fairness criteria for a fair distribution of the rewards: (1) contribution fairness: the data owner's payoff should be positively related to its contribution, (2) regret distribution fairness: the difference of regret among data owners should be minimized and (3) expectation fairness: the variability of data owners' regret should be minimized. Here, regret refers to the difference between the reward already received by the data owner and what they are supposed to receive. 

    \item \textbf{Participation fairness:} Including under-represented and never-represented clients \cite{shi_towards_2024} in the training process. Client selection in FL \cite{cho2022towards} is among the most studied steps of the FL lifecycle, as the subset selection can determine who gets to influence the model and how fast the FL model will converge. Partial participation and uneven representation of client groups might appear to be the easiest aspect to assess. Yet, given the complex and heterogeneous nature of the clients, in addition to the private nature of their data and other properties, it is challenging to determine which properties should guide client selection and how to assess the fairness of this process. The most straightforward approach is to define it at an individual level, where all clients are given an equal chance to contribute. This is often formulated as a long-term fairness constraint \cite{huang_efficiency-boosting_2021}, which requires setting a minimum threshold on the number of times a client needs to be included in the FL process. Another option is to define client groups based on the general knowledge of the system, such as geographical location \cite{lee2024geographical, saputra2019energy} (e.g., applications that might be impacted by the weather or the culture), or language preferences (e.g., next word prediction) \cite{hard2018federated}. 
    
    
\end{itemize}

While these definitions offer technically tractable objectives, they are often introduced in isolation from questions of power, structural inequality, or downstream harm. In the following section, we examine how these fairness notions fall short when applied without regard to the use case context, stakeholders involvement and systemic constraints. 

%% file: 03_failures.tex
\section{Method}
To inform our position, we conducted a systematic literature review covering 121 papers that focus on bias and fairness in FL. We used DBLP to find papers with keywords (`Fair*'+ `Federated learning'), (`Bias'+ `Federated Learning'), which were published up until December 2024. As FL is multidisciplinary, we did not limit our search to ML venues, but also included papers from networking/ telecommunications and economics venues. We filtered out: (a) Comments, abstracts, thesis, tutorials, slides, and other reports which are not research papers, (b) Papers that use of term ``bias'' in other contexts, such as bias-variance tradeoff, (c) duplicates, and duplicates under different titles, (d) papers in predatory venues, and those behind paywalls and inaccessible through the authors' institutions.

\textbf{Categorization:} We manually categorize the papers depending on the fairness definition that was used, i.e., participation fairness, performance fairness, group fairness, and collaborative fairness. Note that many of these papers do not explicitly use this terminology, and that some overlap exists across the definitions. Further filtering was done for papers that explore the intersection with privacy and adversarial robustness, which we considered as a separate category when needed, although they might use the same fairness definitions. 

   \textbf{Annotation:} For each paper, we identified (1) the motivations of the paper and if it came with real-world examples, (2) the fairness definitions and the considered stakeholders, (3) the proposed interventions and which stages of the FL lifecycle they focus on; and (4) the datasets used in the evaluation. 


\section{From Definition to Deployment: Five Pitfalls in Federated Learning Fairness Research}
\label{sec:pitfalls}

Despite increasing attention to fairness in FL, many existing approaches operate with narrow or conflicting definitions, often optimized for system-level performance or developer convenience rather than equity for clients or end users. Drawing from our annotation of 121 papers, we identify five recurring pitfalls that hinder meaningful fair FL research and development. 
\begin{figure}[t]
\centering
\resizebox{0.75\columnwidth}{!}{%
\begin{tikzpicture}
\pie[
    text=legend,
    radius=2.0,
    color={orange!30, blue!40, red!60, green!50, gray!50}
]{ 
    8.4/Participation fairness,
    28.0/Performance-centered,
    34.0/Group fairness-inspired,
    21.6/Collaborative fairness,
    8.0/Surveys and other
}
\end{tikzpicture}
}
\caption{Fairness paradigms covered in our analysis.}
\label{fig:fairness-pie}
\end{figure}
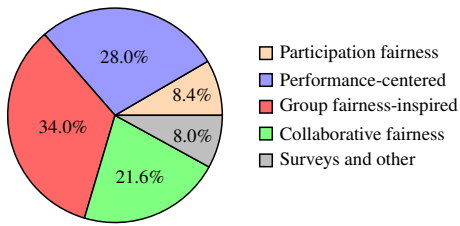



\subsection{Abstract System Formulation: Narrow Focus on Client-Server Architecture}
One common observation is that fairness is FL is defined without explicit articulation of \textit{who is being protected}, \textit{what harms are being mitigated}, or \textit{whose values are being encoded}. Definitions are often not related to stakeholder groups and contextual risks and harms.
Several papers define fairness objectives without specifying whether they are optimizing for institutional clients or individual users, nor do they specify in which use-cases their approaches would be applicable. 

FL is often described as a two-party protocol between a central server and distributed clients. This technical framing---rooted in system architecture---focuses on communication efficiency \cite{lim2020federated}, model convergence \cite{lu2024federated}, and privacy guarantees \cite{yin2021comprehensive} between these two roles. However, real-world FL deployments rarely involve just servers and clients. Instead, they are embedded in larger sociotechnical systems with multiple stakeholders, each with distinct roles, interests, and vulnerabilities \cite{chehbouni2025enhancingprivacyearlydetection,antunes2022federated}.

For example, in cross-silo FL between healthcare institutions, the “client” may represent an entire hospital, but the downstream impact of the model affects patients, physicians, and administrators. Similarly, in financial applications, in addition to customers on whom decisions apply, regulators and auditors may also have a stake in how models are trained and evaluated. Developers, platform providers, and even the designers of benchmark datasets exert influence over what is optimized, who is included, and what trade-offs are made.

This framing risks can lead to what \cite{selbst2019fairness} call an “abstraction error” : treating fairness as a property of a technical subsystem, rather than of the broader sociotechnical system it serves. Justice is not an outcome of optimizing model convergence or performance variance in isolation; it requires attention to several aspects, especially power dynamics and harm.  It requires understanding who is behind each abstract client, who bears the risk when performance degrades, and who gets excluded when design decisions are made. Treating clients as impersonal and theoretical entities misses the nested and the broader institutional, social, and economic factors that shape FL. 
This lack of clarity is not just a technical oversight. It reflects a deeper failure to treat fairness as a \textit{sociotechnical} concept in this subset of the research community. 

\subsection{Abstract Evaluation: Synthetic Clients vs Real Harms}

There is a significant mismatch between the concept that researchers aim to measure (fairness) and the measurement methods they use to evaluate it. This mismatch appears in two ways: the creation of synthetic clients focused solely on data heterogeneity, and the use of datasets unrelated to the motivating use-cases.


The majority of the annotated papers evaluate their framework using centralized ML datasets (e.g., MNIST, CIFAR-10, Adult, or COMPAS), by splitting them into synthetic ``clients.'' 
For instance, for group fairness papers, many studies cite real-world applications in finance, justice, and healthcare, but evaluation typically relies on centralized ML datasets like Adult (\cite{kohavi1996uci}), COMPAS (\cite{angwin2016machine}), and CelebA (\cite{liu2015deep}). A few use the more realistic ACS dataset (\cite{ding_retiring_2021}), which models demographic variation across U.S. states, and only one paper both motivates and evaluates using real-world federated data (\cite{chu2024multi}). 
While useful for prototyping, this practice obscures the sociotechnical context in which fairness actually matters.  This is particularly concerning given that federated learning introduces unique challenges not present in centralized settings (\S~\ref{subsec:fairnesschallenges}).

Only a small fraction of the literature uses datasets from real-world federated settings or evaluate across realistic data from silos or clients. In our analysis, we found that fewer than 10\% of papers evaluated their fairness methods on domain-specific data. 

For instance, in \textit{performance fairness} papers, healthcare is the most commonly cited motivating use case, with the goal of reducing disparities across silos (e.g., hospitals) \cite{yan_new_2024,xu_achieving_2022}. Similar concerns apply to cross-device FL, especially in personalized applications like mobile phones \cite{hard2018federated}, where QoS must be consistent across clients. Yet, most research is evaluated on ML benchmarks like MNIST and CIFAR10, simulating heterogeneity through class imbalance or rotated images. These datasets do not capture real-world distributional shifts, which are shaped by demographic, geographic, or temporal factors, nor do they capture all the domain-specific challenges that come with the application.
Instead, fairness is reduced to an oversimplified distributional problem, where the minority class serves as a proxy for real-world disadvantaged groups.
Only five performance fairness papers evaluate on domain-specific datasets \cite{yan_new_2024,xia_enhancing_2024,xu_achieving_2022, mashhadi_auditing_2022,selialia_federated_2022}, mostly in healthcare.

Likewise, of the 26 papers on \textit{collaborative fairness}, only five are grounded in real use cases \cite{maheswari_transforming_2024, albaseer_fedpot_2024, xu_reciprocal_2024, donahue_models_2021, lu_toward_2022}, three offer motivating examples \cite{lyu_collaborative_2020, yu_fairness-aware_2020, liu_fairshare_2023}, and just one \cite{albaseer_fedpot_2024} conducts experiments using data relevant to its stated application.

While these setups enable reproducibility and control, they abstract away the institutional and social contexts in which FL systems operate. As the technical interventions for fairness in FL are evaluated only over synthetic clients, it is easy to miss the real impact of these solutions. 

Additionally, clients are modelled as interchangeable units (i.e., similar resources, contexts), as the focus is predominantly on data heterogeneity. The abstraction of equally resourced, equally motivated, or equally impacted, hides some of the inequalities fairness mechanisms are meant to address.

\subsection{Protection for the System, Not the Vulnerable}
Another trend we observed is the prioritization of fairness definitions that aim to protect the system from free riders and adversaries, rather than the people the system serves.
\textit{Collaborative fairness}, which represents almost a third of the annotated papers, allocates rewards based on contribution to global utility, often estimated via Shapley values. While this may seem reasonable in settings with symmetric power (e.g., corporate collaboration), in reality, it easily breaks down, as participation is shaped by unequal access to resources and heterogeneous data and contexts. In such settings, clients with noisy or underrepresented data would be penalized not because their data is intentionally low quality, but because it simply deviates from the majority's data. These frameworks' claimed goal is to provide structured methods for evaluating contributions and distributing rewards, and by doing so, they may inadvertently prioritize individual gains over collective outcomes. This individualistic perspective can conflict with the inherently cooperative nature of FL, where the primary objective is to collaboratively train a high-quality global model that benefits \emph{all} participants. 

This becomes obvious with how collaborative fairness papers assign degraded models or lower rewards to clients deemed “low contributors” \cite{lyu_collaborative_2020, yi_fedpe_2024, adamek_privacy-preserving_2024}, without considering that those clients might face structural barriers to participation.
This logic is particularly troubling in high-stakes domains like healthcare or finance---often cited as motivating use-cases for this fairness definition--- where a poorly performing model may harm patients or customers rather than the institution itself. 
Ultimately, it is the patient or customer who pays the price.

 Although free-riders present a potential risk, tackling such issues aligns more closely with adversarial robustness than algorithmic fairness. Collaborative fairness 
 papers appear to prioritize adversarial robustness under the guise of protecting fairness for “honest” clients. Framing these mechanisms as fairness-preserving obscures the reality that they primarily aim to safeguard the system from misuse. While adversarial robustness frameworks often acknowledge the risk of false positives and incorporate mechanisms for correction \cite{touat_towards_2023}, collaborative fairness schemes frequently cast exclusionary decisions, such as penalizing low-contribution clients, as inherently just. This framing leaves little room to question whether such clients are disadvantaged due to resource constraints rather than bad faith \cite{green2021contestation}.


\subsection{Disconnected interventions: One-Stage Solutions to Lifecycle Harms}
Existing interventions in FL in general, and fairness in FL in particular, tend to concentrate on a limited subset of the development pipeline, most notably aggregation schemes \cite{qi2024model} and client selection \cite{10197174}. However, other critical stages such as problem formulation, model initialization, and evaluation remain largely overlooked. This narrow focus creates blind spots: biases introduced in earlier stages may propagate unchecked, and the criteria used to assess fairness may fail to capture the outcomes that matter most. Our analysis reveals that few papers adopt a lifecycle perspective, which is essential to understanding how harms emerge and to designing effective and equitable interventions. Most interventions are localized on a few steps while others are overlooked, as illustrated in Fig \ref{fig:lifecycle_papers}.

Indeed, participation fairness papers focus on client selection \cite{cho2022towards,huang_efficiency-boosting_2021,javaherian_fedfair3_2024} and asynchronous model aggregation \cite{gao_mitigating_2024, wang_libra_2024}. While a large portion of the surveyed literature in group fairness in FL \cite{ezzeldin_fairfed_2021,makhija_achieving_2024,roy_fairness_2024,yue_gifair-fl_2021,meerza_glocalfair_2024,papadaki_minimax_2022,selialia_mitigating_2023,abay_mitigating_2020,li_fairness-aware_2024} proposed techniques that combines adapting local training (e.g., using regularizers, adversarial debiasing) with another intervention at the server-level during model aggregation (e.g., reweighting). For this line of work, only one intervention was proposed for client selection \cite{zhang_fairfl_2020},  and one at the personalization stage \cite{chu2024multi}. Meanwhile, to achieve performance fairness, the proposed solutions span a larger portion of the FL pipeline compared to group fairness, but the different steps have received varying attention. The interventions include model initialization \cite{chu_rethinking_2024}, client selection \cite{javaherian_fedfairmbox3_2024}, adjusting local training \cite{cong_ada-ffl_2024}, model aggregation \cite{li_fair_2020}, or personalization techniques \cite{xie_accelerating_2024} such as clustered FL \cite{chu_focus_2022}. 
Despite these contributions, the disjointed nature of current interventions leaves key sources of harm unaddressed; without a holistic, lifecycle-oriented approach, fairness efforts risk being reactive, fragmented, and insufficiently adapted to the clients heterogeneity.



\subsection{Disconnected Impact: No Single Fairness Definition is Enough}
In many real-world FL scenarios, fairness cannot be reduced to a single definition. Each of the paradigms discussed (i.e., performance-centric, group fairness, and collaborative fairness) captures only one axis of concern. Treating them as mutually exclusive leads to incomplete or even counterproductive interventions. In our analysis, few papers considered multiple fairness at once \cite{amiri_impact_2022,galvez_enforcing_2021}, however, these papers were mostly exploring the impact of privacy techniques on performance and group fairness, rather than their interplay. Instead, we believe a holistic evaluation for different fairness aspects should be adopted. FL is a multi-stakeholder system, thus, it is important to evaluate the harm to each stakeholder. 


Consider a cross-silo FL deployment in the financial sector; one of the few recurring examples in the three different parts of the annotated literature. Multiple banks may collaborate to train a shared fraud detection model. A performance-centric definition might require that the global model perform well for all participating institutions. However, each institution may serve different customer bases, raising concerns about group fairness: will the model perform equally across demographic subgroups within each silo? At the same time, institutions have competing interests. Supposing that the institutions are equally powerful, a so-called collaborative fairness mechanism may be considered to ensure that no institution intentionally degrades the performance for others. In this example, the fairness concerns are not only relevant, they are entangled. Optimizing for just one could undermine the others. For instance, maximizing global performance may disproportionately benefit larger clients; enforcing group fairness locally may reduce global accuracy; prioritizing contribution-based rewards may punish institutions that serve minority populations.

This illustrates a broader point: fairness in FL is not a property of a single metric or stage in the pipeline. It is an emergent property of the full system, shaped by technical design choices, institutional power dynamics, and the social roles of clients and end users. Addressing fairness meaningfully requires acknowledging this multidimensionality/ multi-stakeholder nature, and moving beyond one-size-fits-all definitions.

%% file: 04_bias_framework.tex
\begin{figure}[t]
\centering
\begin{tikzpicture}
\begin{axis}[
    x=1.2cm,
    ybar,
    bar width=0.15cm,
    width=\columnwidth,
    height=5.5cm,
    ymin=0, ymax=25,
    ylabel={\footnotesize \# Papers},
    xtick=data,
    xtick pos=left,
    ytick pos=left,
    tick align=outside,
    xticklabel style={rotate=45, anchor=east, font=\scriptsize},
    symbolic x coords={
        Init/Preproc.,
        Client Sel.,
        Local Train.,
        Aggreg.,
        Eval.,
        Add. Mech.
    },
    nodes near coords,
    nodes near coords style={font=\scriptsize, text=black},
    bar shift=0pt,
    legend style={font=\scriptsize, at={(0.5,-0.35)}, anchor=north, legend columns=2, /tikz/every even column/.style={column sep=0.2cm}},
    title={Lifecycle Steps Targeted per Fairness Paradigm},
    title style={font=\footnotesize},
]

\addlegendimage{empty legend}
\addlegendimage{empty legend}

\addplot+[bar shift=-0.3cm, fill=blue!40, legend image code/.code={
        \draw[fill=blue!40] (0cm,-0.1cm) rectangle (0.3cm,0.1cm);
    }] coordinates {
    (Init/Preproc.,2)
    (Client Sel.,8)
    (Local Train.,6)
    (Aggreg.,16)
    (Eval.,8)
    (Add. Mech.,2)
};

\addplot+[bar shift=-0.1cm, fill=red!60, draw=black, legend image code/.code={
        \draw[draw=black, fill=red!60] (0cm,-0.1cm) rectangle (0.3cm,0.1cm);
    }] coordinates {
    (Init/Preproc.,2)
    (Client Sel.,1)
    (Local Train.,22)
    (Aggreg.,17)
    (Eval.,1)
    (Add. Mech.,7)
};

\addplot+[bar shift=0.1cm, fill=green!50, legend image code/.code={
        \draw[fill=green!50] (0cm,-0.1cm) rectangle (0.3cm,0.1cm);
    }] coordinates {
    (Init/Preproc.,0)
    (Client Sel.,3)
    (Local Train.,0)
    (Aggreg.,9)
    (Eval.,0)
    (Add. Mech.,5)
};

\addplot+[bar shift=0.3cm, fill=orange!30, legend image code/.code={
        \draw[fill=orange!30] (0cm,-0.1cm) rectangle (0.3cm,0.1cm);
    }] coordinates {
    (Init/Preproc.,0)
    (Client Sel.,8)
    (Local Train.,1)
    (Aggreg.,2)
    (Eval.,0)
    (Add. Mech.,0)
};

\legend{\hspace{0.5cm}\textbf{Fairness}, \hspace{-1.4cm}\textbf{Paradigm}, Performance, \hspace{-0.6cm}Group, Collaborative, Participation}

\end{axis}
\end{tikzpicture}
\caption{Lifecycle steps addressed in fairness papers, by fairness paradigm.}
\label{fig:lifecycle_papers}
\end{figure}
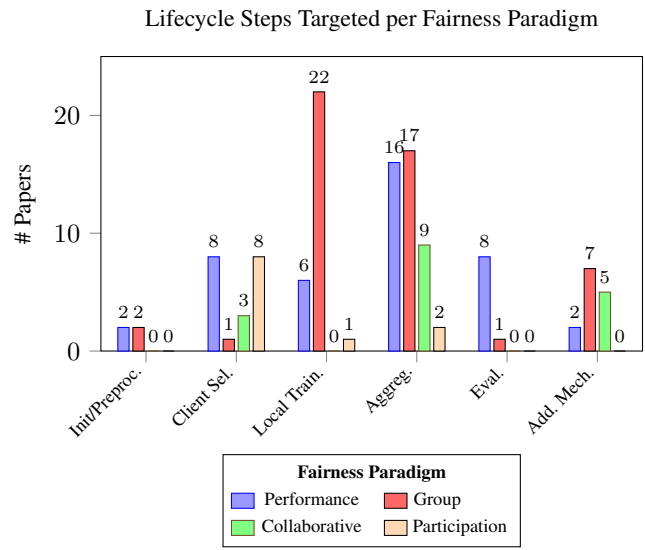

\begin{figure*}[t]
    \centering
\includegraphics[width=0.99\linewidth]{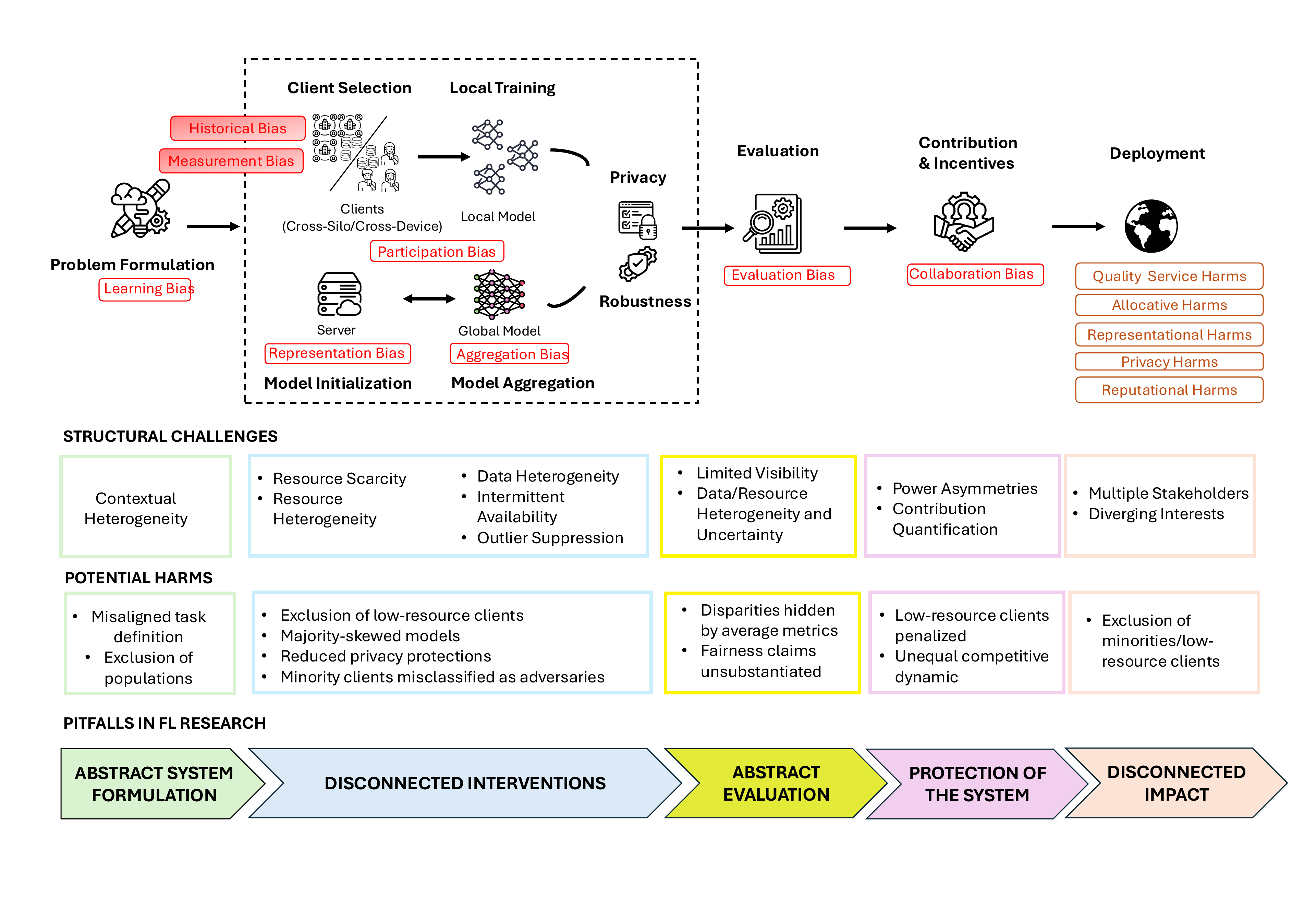}
    \caption{Overview of our harm-centered framework for fairness in FL. At each step of the FL lifecycle, we identify biases within stakeholders control, or outside of their control (red-filled boxes), structural challenges and potential harms that might emerge at each step.}
    \label{fig:harmtaxo}
\end{figure*}

\section{A Harm-Centered Framework for Fairness in Federated Learning}

ML systems can lead to harms such as unequal performance across demographic groups, inequitable resource distribution, or the reinforcement of stereotypes \cite{shelby2023sociotechnical}. FL inherits these risks from traditional ML while introducing new ones due to its decentralized structure and structural asymmetries across clients. To design fairer FL systems, we must shift from identifying isolated biases to understanding how harm emerges from developer decisions throughout the lifecycle, influenced by \textit{heterogeneity} (in data, context, resources) and \textit{scarcity} (of data and compute power).

Following our analysis in Section~\ref{sec:pitfalls}, we propose a framework for identifying potential harms and guiding fairness interventions throughout the FL pipeline. Rather than viewing bias as a fixed category, we examine how developer choices, constrained by heterogeneity across clients, may introduce harms such as quality of service (QoS), allocative, and representational harms. Below, we walk through each stage of the FL lifecycle and highlight where structural conditions and technical design interact to produce or amplify harm. We illustrate our framework in Figure \ref{fig:harmtaxo}.



\subsection{Federated Learning Lifecycle Decisions and Harms}
\label{subsec:lifecycle_harms}

Developers have little control and visibility on data collection and local model training, in comparison with centralized settings. However, different decisions that are made throughout the FL lifecycle propagate existing biases and introduce new ones. Indeed, data generation and collection introduces 1) \textbf{historical bias}: Since data reflects the cultural, institutional, political, and socioeconomic context surrounding its extraction, it can encompass past or existing structural inequalities and injustices (e.g., COMPAS dataset ~\cite{angwin2016machine}); 2)  
\textbf{representation/ selection bias} where these biases arise when training data is not representative of the real-world population or use-case distribution (e.g., ImageNet data \cite{yang2020towards}); and 3) \textbf{measurement bias}: Such bias is introduced by how features or labels are collected, chosen, and used (e.g., faulty sensors, inadequate proxies \cite{kleinberg2018algorithmic}). 
Such biases can be amplified during local training, introducing \textbf{learning bias}. In the following, we focus on the different steps where there is more control and visibility, as well as biases that are specifically tied to FL's collaborative nature.

\paragraph{Problem Formulation} Developers often retain full control over how problems are framed, including defining learning objectives, inputs and outputs, and success metrics \cite{passi_problem_2019,suresh_framework_2021}. In FL, contextual heterogeneity across clients, such as differing institutional goals or regional constraints, complicates this step. When problem formulations assume a universal objective, they may be misaligned with local needs \cite{raji_fallacy_2022}. For instance, health prediction tasks that ignore regional disease prevalence or local medical practices risk producing models that systematically underperform for some populations \cite{asiedu_case_2024}. Additionally, oversimplified fairness criteria may be selected for convenience rather than appropriateness \cite{simson2024lazy}, leading to \textbf{learning bias} \cite{suresh_framework_2021}.
Thus, rather than just asking technical questions about the solution, moving away from the client-server abstraction is necessary. Considering \textit{who this model is for?} \textit{what goals are embedded in this task?} and \textit{whose outcomes matter?} can positively impact the formulation to be more applicable.


\paragraph{Model Initialization} Model choices, including size, architecture, and pretraining, are often made before the start of the training, and in tandem with problem formulation. However, these decisions can exclude clients with limited resources, especially in cross-device FL, where large models may not run on older or lower-powered devices \cite{mcmahan2017communication, bonawitz2019towards}. This creates \textbf{participation bias}, where only well-resourced clients can meaningfully contribute to training. Similarly, using a pre-trained model may introduce \textbf{representation bias} if the pretraining dataset does not reflect the data distributions of FL clients \cite{wang_research_2023, chuang2023debiasing, kaur_comprehensive_2023}.
As a result, it is important to consider the possibility that \textit{the chosen model architecture/ pretrained model disadvantage clients with limited resources or underrepresented data.}

\paragraph{Client Selection} While many aspects about data collection are not under the developers' control, especially in cross-device FL, some of the biases and harms found in data collection are shifted to client selection. 

In cross-device FL, the ideal scenario would include a full participation from all clients at each communication round. Nonetheless, clients are often offline, have low battery/ poor connectivity, or are unavailable \cite{9220170}. Additionally, privacy concerns in FL due to memorization remain present \cite{thakkar2020understanding}. If clients participate too often, their data may be over-exposed during model updates. This leads to the ideal to be purely random/ round robin selection. Moreover, the most adopted FL aggregation scheme is synchronous, meaning that if a client does not send their update before a set deadline, their update is discarded.  Additionally, the communication bandwidth is a bottleneck for FL, hence only a few clients can participate in each round. Thus, optimizing the client selection step  has become a key strategy in FL 
\cite{nishio2019client, zhu2022online, cho2022towards}. The proposed schemes often try to maximize the number of collected updates in order to improve the performance of the global model. While these schemes might accelerate the convergence of the training on average, these schemes are often biased towards clients with more powerful devices and better connectivity. Slow clients are often labeled as stragglers and dropped out of the training round.
On the opposite side of the spectrum, other strategies focus on data properties, where clients are selected in proportion to the size or quality of their data, meaning that clients with more data (or potentially more valuable data) are more likely to be selected \cite{goetz_active_2019}. However, data size is not a good proxy for the meaningfulness of a contribution due to redundancy \cite{9220170}. Moreover, clients with larger datasets (e.g., big companies or institutions with more resources) may dominate the model, potentially drowning out smaller clients or those with less data but more diverse or unique insights. 

Client selection strategies that overlook resource and data heterogeneity and their interplay are how \textbf{participation bias} could manifest in FL. Consequently, it is important to ask \textit{Are some clients less likely to participate? Are selection rules exclusionary?}


\paragraph{Model Aggregation} FL aggregation combines updates from clients to form a global model. There are two decisions that developers make for this stage: \textit{When to aggregate?} and \textit{How to weigh the updates?} For the first question, synchronous aggregation is the most commonly adopted scheme. It requires that model updates are averaged or combined at fixed intervals, with updates being sent after the fixed deadlines being ignored. This inevitably exacerbates \textbf{participation bias}, as clients with slower internet connections or lower computational power might not be able to update their models in time for synchronization \cite{gao_mitigating_2024}. Over time, this would lead to a model that better represents the data from faster clients but fails to accurately capture the diversity and variability of the entire population of clients.

The second decision is choosing the updates weighting scheme. The first proposed and most adopted method \cite{mcmahan2017communication} requires weighting models based on the size of each client’s dataset. Such technique amplifies aggregation bias by prioritizing dominant clients and suppressing minority contributions. Over time, this leads to models that reflect majority distributions and perform poorly for smaller, less represented clients, leading to \textbf{aggregation bias}. Different aggregation schemes were proposed \cite{qi2024model}, tackling various challenges in FL, including robustness and asynchronous updates. This step is also a popular target of group fairness in FL techniques.\\ 
As a decisive step, it is important to ask \textit{1) How does the aggregation deadline affect which clients are able to participate, and which are excluded?}, and \textit{2) How does the aggregation weighting scheme influence the fairness outcome being targeted?}

\paragraph{Evaluation} 
 FL evaluations often rely on benchmark datasets or average client performance. However, due to limited visibility into local data, these metrics may obscure disparities \cite{lai2022fedscale}. Similarly to centralized ML, this can be referred to as \textbf{evaluation bias}, where the methods or conditions used to assess a model’s performance do not accurately represent the real-world scenarios in which it will be deployed. A model that performs well on average may still harm specific clients, especially if fairness is not evaluated across subgroups. Additionally, relying solely on benchmarks and simulated environments can further distort conclusions. 
 While these tools offer necessary step for developers to prototype, these remain controlled conditions with standardized datasets and predefined assumptions, whereas real-life conditions would differ significantly. Thus it is necessary to adopt detailed metrics (e.g., performance, client participation and drop-out, group-fairness metrics), and to continually monitor the various aspects of the FL process after deployment. 

For this part we should ask \textit{Which metrics are we using? On which data do we evaluate? For whom do these metrics matter? and how often do we need to assess the system's performance?}

\paragraph{Contribution and Incentives}
Incentive mechanisms in FL aim to reward clients based on their contributions to the global model. However, measuring “contribution” remains an open and contested problem. Existing approaches often favour clients whose data aligns with dominant patterns or leads to measurable performance gains in global utility. Meanwhile, clients providing rare or minority data, which may be critical for generalization and group or performance fairness, are undervalued or excluded. This creates a feedback loop that reinforces inequality and marginalizes those already underrepresented, we refer to this issue as \textbf{collaboration bias}.

In FL, clients already bear real costs—computational, communicational, and sometimes organizational. In theory, participation should offer sufficient returns to justify these investments, whether through improved local model performance, access to the global model, or monetary compensation.
While improving contribution evaluation is a meaningful technical objective, it also risks missing the larger question: \textit{What purpose do additional incentives serve in the first place?} To move forward, future work must grapple with contextual questions: \textit{Who is affected by the final model? What are the actual costs borne by clients? How would an incentive scheme shape participation and power within the federation?} Without answers to these questions, incentive design risks introducing new harms—through \textbf{collaborative bias} where incentive mechanisms introduce misattributed rewards and the punishment of honest but nonconforming clients.

\paragraph{Privacy and Robustness Mechanisms} Privacy guarantees and robustness defences are essential in FL, but they are not neutral. 

\textit{Privacy:} As keeping the data local was proven not enough for privacy protection, a combination of FL with privacy preserving mechanisms such as differential privacy (DP) and secure aggregation have been adopted in the literature and practice through the use of DP-SGD for instance.
However, balancing privacy and fairness presents a challenge. On one hand, DP degrades the performance of the models, and disproportionately affect underrepresented groups \cite{bagdasaryan_differential_2019, ling_fedfdp_2024}. On the other hand, prioritizing fairness can heighten privacy risks due to revealing sensitive attributes and memorization \cite{pentyala_privfairfl_2022}. Given these tradeoffs and disparate impacts on privacy and performance, it is important to consider
\textit{what are the side effects of the privacy mechanisms?} and ask if \textit{everyone is equally protected?}

\textit{Adversarial Robustness:} Adversarial attacks in FL, such as model/data poisoning, free riding, and backdoor attacks, are often detected through anomaly detection or using similarity metrics, using techniques like reputation systems \citet{lyu_collaborative_2020}, performance evaluation, and statistical tests (e.g., clustering or outlier detection) \cite{sattler2020clustered}. However, due to data heterogeneity, falsely identifying a legitimate client as an adversary is a common risk with significant consequences \cite{touat_towards_2023, ren_cosper_2024}. Misclassified clients may face unjust reputation damage, exclusion from future tasks, financial losses tied to missed incentives, and reduced opportunities for collaboration. These outcomes can erode trust, foster alienation, and ultimately discourage participation in FL.
When designing robustness mechanisms, developers must ask: \textit{What are the consequences of a false-positive detection—and is the cost of excluding a legitimate client greater than the risk posed by an undetected adversary?}
It is thus necessary to ensure that these additional mechanisms \textit{do not introduce additional harms}, while providing \textit{equal protection} to different stakeholders.

\subsection{Downstream Harms in Federated Learning}

FL inherits many of the fairness concerns from centralized ML but introduces new layers of complexity due to its distributed nature and client heterogeneity. In this section, we examine different types of harms that can arise in FL, and we map each to the corresponding biases and stages in the FL lifecycle that contribute to them. Understanding how harms emerge and propagate is essential for developing targeted and context-aware fairness interventions.

In Section \ref{subsec:lifecycle_harms}, we noted that several decisions lead to the exclusion of minority populations. Such harms in practice translate into QoS, allocative, and representational harms. While such harms are noted in centralized ML, they take different forms in FL settings.  Additionally, additional mechanisms for privacy and adversarial robustness introduce disparities and possible damages.   In particular, the biases in the FL lifecycle translate into the following harms:

\begin{itemize}

\item \textbf{Quality of Service (QoS) Harms.} These occur when the FL model performs unequally across clients. A typical cause is \textit{aggregation bias}, where global model updates disproportionately reflect data from dominant clients (e.g., those with more frequent participation or more standard data distributions). This harm is often introduced during model aggregation \cite{michieli_are_2021}, but can be seeded earlier through biased model initialization or improper problem formulation (e.g., assuming all clients have similar objectives \cite{suresh_framework_2021}). 

\item \textbf{Allocative Harms.} These relate to unfair allocation of benefits or resources informed by FL models. They may result from \textit{representation bias} or \textit{learning bias} at the data and training stages, where certain groups are underrepresented or oversimplified. In financial applications, for instance, credit scoring systems built using FL can exclude certain regions or demographics if their data was sparsely sampled or downweighted \cite{ding_retiring_2021}. Incentive mechanisms that reward ``high-contributing'' clients also risk allocative harms when contributions are unfairly evaluated \cite{albaseer_fedpot_2024}.

\item \textbf{Representational Harms.} FL systems may reinforce stereotypes or marginalize specific groups when measurement or historical bias is present at the data collection step \cite{yang2020towards, kleinberg2018algorithmic}. These harms are especially prominent in applications involving vision \cite{li2024rectify} language \cite{gallegos2024bias}. 

\item \textbf{Privacy Harms.} While FL aims to preserve privacy by design, privacy-enhancing techniques like differential privacy can have disparate effects. For instance, clients with small datasets or outlier data distributions may suffer from higher performance degradation due to added noise \cite{bagdasaryan_differential_2019}. These harms are introduced by protection mechanisms, but are shaped by earlier choices around data representation and client sampling.

\item \textbf{Reputational Harms.} When robustness mechanisms such as anomaly detection or reputation scoring are deployed, clients with nonconforming updates (e.g., due to non-IID data) risk being misclassified as adversaries \cite{touat_towards_2023, ren_cosper_2024}. This can result in exclusion from training, lost incentives, or damaged reputations, even when clients acted in good faith. These harms often emerge during aggregation or incentive distribution stages but stem from unaccounted-for data heterogeneity.

\end{itemize}

%% file: 05_takeaways.tex
\section{Takeaways and Path Forward}

Our analysis of fairness in FL highlights a disconnect between how fairness is defined in research and how harms manifest in practice. If fairness is to be meaningful in FL, it must be grounded not only in optimization objectives, but in the structural conditions and institutional realities shaping the collaboration.

We propose a framework emphasizing that fairness in FL is not a single technical problem, but a distributed, sequential, multi-stakeholder and context-dependent challenge. Addressing fairness requires understanding where structural inequalities shape developer decisions, and how those decisions affect different stakeholders in different ways.
While we focused on the classic horizontal FL lifecycle, it is important to recognize that the pipeline is not fixed—additional steps such as personalization or clustering may be introduced, and design choices at each stage can evolve to better address fairness and contextual needs.

Following our analysis, We note some key takeaways for future work: 
\begin{enumerate}
    \item \textbf{Fairness requires contextually grounded evaluation.} The datasets most commonly used in FL fairness research are inherited from centralized ML and fail to reflect the complexity of collaborative scenarios. Many rely on synthetic or oversimplified distributions, with minority classes used as proxies for structural disadvantage. These abstractions risk obscuring real harms and may result in cherry-picked evaluations \cite{lones2021avoid}. 
    Only a few datasets, such as folktables \cite{ding_retiring_2021} and flamby \cite{ogier2022flamby}, reflect real-world fairness/ FL concerns, yet they remain limited in scope. Advancing fairness in FL will require better datasets, more realistic simulations, and evaluation strategies that reflect real-world risks.

    \item \textbf{Fairness must account for real-world harms and stakeholder impact.} Understanding fairness in FL requires identifying who is affected, by which kinds of errors, and how these translate into tangible harms. This includes underrepresented clients misclassified as adversaries \cite{touat_towards_2023, ren_cosper_2024}, institutions excluded due to limited data \cite{goetz_active_2019}, or regions harmed by biased global updates \cite{asiedu_case_2024}. Researchers must go beyond convergence metrics to consider harms like reduced quality of service, loss of trust, and perpetuation of inequity.

    \item \textbf{Lifecycle harms require lifecycle thinking.} Many fairness interventions in FL target isolated steps, with a focus on client selection and aggregation. However, harms often emerge from a chain of design choices across the FL lifecycle. For example, performance disparities may stem as much from problem formulation or model initialization as from aggregation. Tackling fairness effectively thus demands a holistic view: interventions should account for how earlier decisions constrain fairness downstream, and solutions must span multiple stages to achieve the fairness goals.

    \item \textbf{Fairness definitions are not interchangeable.} Performance, group, and collaborative fairness objectives may conflict, especially in heterogeneous systems. In some FL settings—such as financial or healthcare collaborations—multiple fairness notions may be simultaneously relevant. For instance, clients may seek equitable performance (performance fairness), while also serving diverse subpopulations (group fairness), and contributing to shared resources (collaborative fairness). Future work should investigate how these fairness objectives interact and how conflicts between them can be resolved.
    
    \item \textbf{Stakeholders must shape the design process.} Fairness cannot be defined externally to those it affects. Participatory approaches \cite{birhane2022power} that engage clients, domain experts, and impacted communities can improve problem formulation, risk assessment, and modeling decisions. As FL distributes control and responsibility across many actors, it also demands more inclusive governance mechanisms. Harms cannot be addressed without first identifying who could be harmed, which errors are most consequential, and how these manifest in specific contexts \cite{chehbouni2025enhancingprivacyearlydetection, shelby2023sociotechnical, abercrombie2024collaborative}.
\end{enumerate}

%% file: 06_limitations_idk.tex
\section{Limitations}
Our analysis is subject to several limitations. First, we focus primarily on horizontal FL, as it is the most widely studied variant in fairness literature; however, other forms such as vertical FL present different challenges that merit separate examination. 
Second, the coverage of our annotations may be constrained by the chosen library and search criteria, although we believe the sample is representative of the current research landscape. Finally, despite our effort to systematize the annotations, the process inevitably involves interpretation, which may introduce subjectivity or overlook subtle nuances. 

